\def\year{2020}\relax
\documentclass[letterpaper]{article} 
\usepackage{aaai20}  
\usepackage{times}  
\usepackage{helvet} 
\usepackage{courier}  
\usepackage[hyphens]{url}  
\usepackage{graphicx} 
\usepackage{graphicx}  
\usepackage{epsfig}
\usepackage{amsmath}
\usepackage{amssymb}
\usepackage{multirow}
\usepackage{booktabs}
\usepackage{multicol}
\usepackage{subfigure}
\usepackage{color,xcolor}
\usepackage{colortbl}
\nocopyright

\title{Progressive Feature Polishing Network for Salient Object Detection}
\author{
\Large \textbf{Bo Wang,\textsuperscript{\rm 1,2} Quan Chen,\textsuperscript{\rm 2} Min Zhou,\textsuperscript{\rm 2} Zhiqiang Zhang,\textsuperscript{\rm 2} Xiaogang Jin,\textsuperscript{\rm 1} Kun Gai\textsuperscript{\rm 2}}\\ 
\Large\textsuperscript{1}Zhejiang University,\textsuperscript{2}Alibaba Group\\
wangbo060@zju.edu.cn,\{chenquan.cq,yunqi.zm,zhang.zhiqiang,jingshi.gk\}@alibaba-inc.com,jin@cad.zju.edu.cn 
}
 
\begin{document}

\maketitle

\begin{abstract}
	Feature matters for salient object detection.
	Existing methods mainly focus on designing a sophisticated structure to incorporate multi-level features and filter out cluttered features.
	We present Progressive Feature Polishing Network~(PFPN), a simple yet effective framework to progressively polish the multi-level features to be more accurate and representative.
	By employing multiple Feature Polishing Modules~(FPMs) in a recurrent manner, our approach is able to detect salient objects with fine details without any post-processing.
	A FPM parallelly updates the features of each level by directly incorporating all higher level context information.
	Moreover, it can keep the dimensions and hierarchical structures of the feature maps, which makes it flexible to be integrated with any CNN-based models.
	Empirical experiments show that our results are monotonically getting better with increasing number of FPMs.
	Without bells and whistles, PFPN outperforms the state-of-the-art methods significantly on five benchmark datasets under various evaluation metrics.

\end{abstract}

\section{Introduction}

Salient object detection, which aims to extract the most attractive regions in an image, is widely used in computer vision tasks, 
including video compression~\cite{guo2010novel},
visual tracking~\cite{borji2012adaptive}, and image retrieval~\cite{cheng2017intelligent}.

Benefitting from the hierarchical structure of CNN, deep models can extract multi-level features that contain both low-level local details and high-level global semantics.
To make use of detailed and semantic information, a straightforward integration of the multi-level context information with concatenation or element-wise addition of different level features can be applied.
However, as the features can be cluttered and inaccurate at some levels, this kind of simple feature integrations tends to get suboptimal results.
Therefore, most recent attractive progress focuses on designing a sophisticated integration of these multi-level features.
We point out the drawbacks of current methods in three folds.
First, many methods~\cite{zhang2018progressive,liu2018picanet} employ the U-Net~\cite{ronneberger2015u} like structure in which the information flow from high level to low level during feature aggregation, while BMPM~\cite{zhang2018bi} uses a bidirectional message passing between consecutive levels to incorporate semantic concepts and fine details.
However, these integrations, performed indirectly among multi-level features, may be deficient because of the incurred long-term dependency problem~\cite{bengio1994learning}.
Second, other works~\cite{zhuge2019def,zhang2018bi,hou2017deeply} recursively refine the predicted results in a deep-to-shallow manner to supplement details.
However, predicted saliency maps have lost the rich information and the capability of refinement is limited.
Furthermore, although valuable human priors can be introduced by designing sophisticated structures to incorporate multi-level features, this process can be complicated and the structure might lack generality.

\begin{figure}[t]
	\centering
	\setlength{\tabcolsep}{1pt}
	\renewcommand{\arraystretch}{0.5} 
	\begin{tabular}{ccc}
	\includegraphics[width=0.32\linewidth,height=0.32\linewidth]{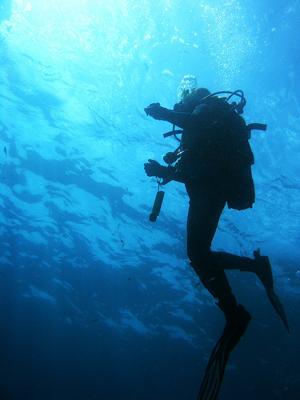}
	&\includegraphics[width=0.32\linewidth,height=0.32\linewidth]{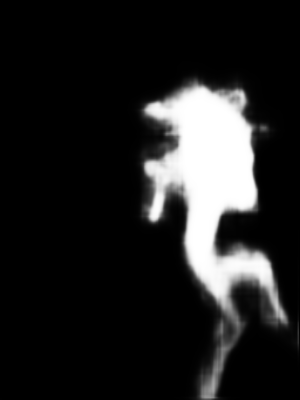}
	&\includegraphics[width=0.32\linewidth,height=0.32\linewidth]{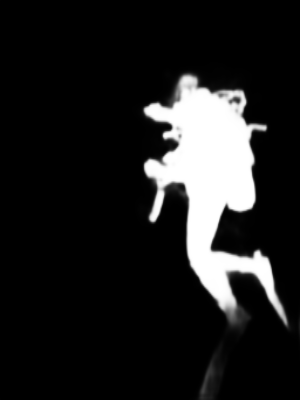}\\
 
	(a) & (b) & (c)\\
 
	\includegraphics[width=0.32\linewidth,height=0.32\linewidth]{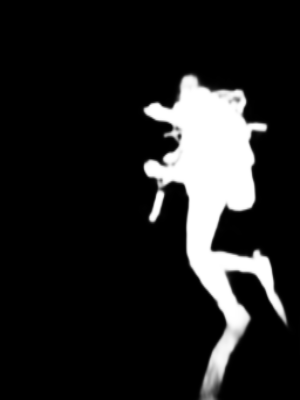}
	&\includegraphics[width=0.32\linewidth,height=0.32\linewidth]{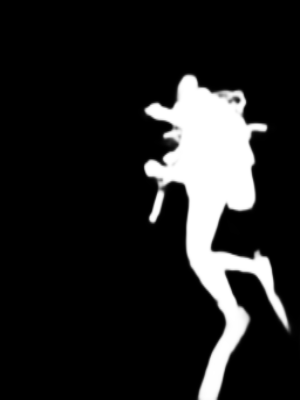}
	&\includegraphics[width=0.32\linewidth,height=0.32\linewidth]{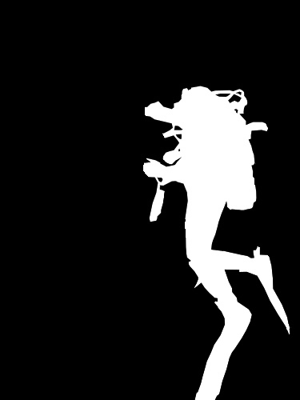}\\
 
	(d) & (e) & (f)\\
	
	\end{tabular}
 \caption{
	 \small Illustration of results with progressively polished features.
	 (a) Original images. (f) Ground truth. (b)-(e)  Saliency maps predicted by PFPN with $T= 0\sim3$ FPMs, respectively.}
 \label{fig:pfp_results}
 \end{figure}

To make full use of semantic and detailed information, we present a novel Progressive Feature Polishing Network~(PFPN) for salient object detection,
which is simple and tidy, yet effective. 
First, PFPN adopts a recurrent manner to progressively polish every level features in parallel.
With the gradually polishing, cluttered information would be dropt out and multi-level features would be rectified.
As this parallel structure could keep the feature levels in backbone, some common decoder structures can be easily applied.
In one feature polishing step, each level feature is updated with the fusion of all deeper level features directly.
Therefore, high level semantic information can be integrated directly to all low level features to avoid the long-term dependency problem.
In summary, the progressive feature polishing network greatly improves the multi-level representations, and even with the simplest concatenation feature fusion, PFPN works well to detect salient objects accurately.
Our contributions are as follows:

$\bullet$
We propose a novel multi-level representation refinement method for salient object detection, as well as a simple and tidy framework PFPN to progressively polish the features in a recurrent manner.

$\bullet$
For each polishing step, we propose the FPM to refine the representations, which preserves the dimensions and hierarchical structure of the feature maps.
It integrates high level semantic information directly to all low level features to avoid the long-term dependency problem.

$\bullet$
Empirical evaluations show that our proposed method significantly outperforms state-of-the-art methods on five benchmark datasets under various evaluation metrics.

\section{Related Work}



During the past decades, salient object detection has experienced continuous innovation.
In earlier years, saliency prediction methods~\cite{itti1998model,parkhurst2002modeling} mainly focus on heuristic saliency priors and low-level handcrafted features, such as center prior, boundary prior, and color contrast.

In recent years, deep convolutional networks have achieved impressive results in various computer vision tasks and also been introduced to salient object detection.
Early attempts of deep saliency models include Li~\cite{li2015visual} which exploits multi-scale CNN contextual features to predict the saliency of each image segment, and Zhao~\cite{zhao2015saliency} which utilizes both the local and global context to score each superpixel.
While these methods achieve obvious improvements over handcrafted methods, their scoring one image patch with the same saliency prediction drops the spatial information and results in low prediction resolution.
To solve this problem, many methods based on Fully Convolutional Network~\cite{long2015fully} are proposed to generate pixel-wise saliency.
Roughly speaking, these methods can be categorized into two lines.

\begin{figure*}[hbt]
	\centering\includegraphics[width=0.98\linewidth,height=0.355\linewidth]{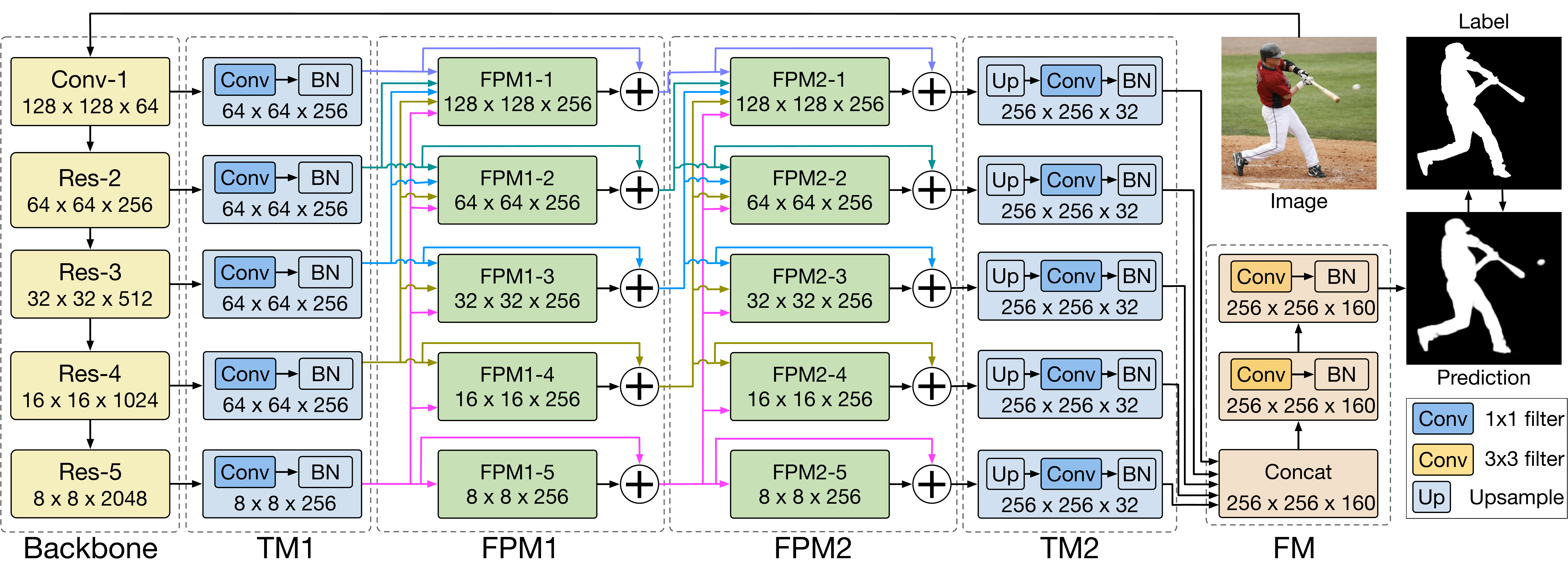}
	\caption{
		\small Overview of the proposed Progressive Feature Polishing Network~(PFPN).
		PFPN is a deep fully convolutional network composed of four kinds of modules: the Backbone, two Transition Modules~(TM), a series of $T$ Feature Polishing Modules~(FPM) and a Fusion Module~(FM). An implementation with ResNet-101~\cite{he2016deep} as backbone and $T=2$ is illustrated. For an input image with the size of 256x256, the multi-level features are first extracted by the backbone and transformed to same dimension by the TM1. Then the features are progressively polished by passing through the two FPMs. Finally, they are upsampled to the same size by TM2 and concatenated to locate the salient objects in FM.
		}
	\label{pipeline}
\end{figure*}

\subsection{Feature Integration}
Although multi-level features extracted by CNN contain rich information about both high level semantics and low level details, 
the reduced spatial feature resolution and the likely inaccuracy at some feature levels make it an active line of work to design sophisticated feature integration structures.
Lin~\cite{lin2017refinenet} adotps RefineNet to gradually merge high-level and low-level features from backbone in bottom-up method.
Wang~\cite{wang2018detect} propose to better localize salient objects by exploiting contextual information with attentive mechanism.
Zhuge~\cite{zhuge2019def} employ a structure which embeds prior information to generate attentive features and filter out cluttered information.
Different from above methods which design sophisticated structure to make information fusion,
we use simple structure to polish multi-level features in recurrent manner and in parallel.
Meanwhile, the multi-level structure would be kept and the polished multi-level features can be applied in common decoder modules.
Zhang~\cite{zhang2018bi} use a bidirectional message passing between consecutive levels to incorporate semantic concepts and fine details.
However, the incorporating the features in between adjacent feature levels results in long-term dependency.
Our method directly aggregates features of all higher levels at each polishing step and thus high level information could be fused to lower level features sufficiently during multiple steps.

\subsection{Refinement on saliency map }
Another line focuses on progressively refining the predicted saliency map by rectifying previous errors.
DHSNet~\cite{liu2016dhsnet} first learns a coarse global prediction and then progressively refines the details of saliency map by integrating local context features.
Wang~\cite{wang2016saliency} propose to recurrently apply an encoder-decoder structure to previous predicted saliency map to perform refinement.
DSS~\cite{hou2017deeply} adotps short connections to make progressive refining on saliency maps.
CCNN~\cite{tang2019salient} cascads local saliency refiner to refines the details from initial predicted salient map.
However, since the predicted results have severe information loss than original representations, the refinement might be deficient.
Different from these methods, our approach progressively improves the multi-level representations in a recurrent manner instead of attempting to rectify the predicted results.
Besides, most previous refinements are performed in a deep-to-shallow manner, in which at each step only the features specific to that step are exploited.
In contrast to that, our method polishes the representations at every level with multi-level context information at each step.
Moreover, many methods utilize an extra refinement module, either as a part of their model or as a post-process, to further recover the details of the predicted results, such as DenseCRF~\cite{hou2017deeply,liu2018picanet}, BRN~\cite{wang2018detect} and GFRN~\cite{zhuge2019def}.
In contrast, our method delivers superior performance without such modules.


\section{Approach} \label{approach}
In this section, we first describe the architecture overview of the proposed Progressive Feature Polishing Network~(PFPN).
Then we detail the structure of the Feature Polishing Module~(FPM) and the design of feature fusion module.
Finally we present some implementation details.

\subsection{Overview of PFPN}
\label{approach:overview}
In this work, we propose the Progressive Feature Polishing Network (PFPN) for salient object detection.
An overview of this architecture is shown in Fig.~\ref{pipeline}.
Our model consists of four kinds of modules: the Backbone, two Transition Modules~(TM), a series of $T$ Feature Polishing Modules~(FPM), and a Fusion Module~(FM).

The input image is first fed into the backbone network to extract multi-scale features.
The choice of backbone structure is flexible and ResNet-101~\cite{he2016deep} is used in the paper to be consistent with previous work~\cite{zhuge2019def}.
Results of VGG-16~\cite{simonyan2014very} version is also reported in experiments.
Specifically, the ResNet-101~\cite{he2016deep} network can be grouped into five blocks by a serial of downsampling operations with a stride of 2.
The outputs of these blocks are used as the multi-level feature maps:  \textit{Conv-1}, \textit{Res-2}, \textit{Res-3}, \textit{Res-4}, \textit{Res-5}.
To reduce feature dimensions and keep the implementation tidy, these feature maps are passed through the first transition module~(TM1 in Fig.~\ref{pipeline}), in which the features at each level are transformed in parallel into a same number of dimensions, such as 256 in our implementation, by 1x1 convolutions.
After obtaining the multi-level feature maps with the same dimension, a series of $T$ Feature Polishing Modules~(FPM) are performed on these features successively to improve them progressively.
Fig.~\ref{pipeline} shows an example with $T=2$.
In each FPM, high level features are directly introduced to all low level features to improve them, which is efficient and notably reduces information loss than indirect ways.
The inputs and outputs of FPM have the same dimensions and all FPMs share the same network structure.
We use different parameters for each FPM in expectation that they could learn to focus on more and more refined details gradually.
Experiments show that the model with $T=2$ outperforms the state-of-the-art and also has a fast speed of 20 fps, while the accuracy of saliency predictions converges at $T=3$ with marginal improvements over $T=2$.
Then we exploit the second transition module~(TM2 in Fig.~\ref{pipeline}), which consists of a bilinear upsampling followed by a 1x1 convolution, to interpolate all features to the original input resolution and reduce the dimension of them to 32.
At last, a fusion module~(FM) is used to integrate the multi-scale features and obtain the final saliency map.
Owing to the more accurate representations after FPMs, the FM is implemented with a simple concatenation strategy.
Our network is trained in an end-to-end manner.

\subsection{Feature Polishing Module}
\label{approach:fpm}
The Feature Polishing Module~(FPM) plays a core role in our proposed PFPN.
FPM is a simple yet effective module that can be incorporated with any deep convolutional backbones to polish the feature representation.
It keeps the multi-level structure of the representations generated by CNNs, such as the backbone or preceding FPM, and learns to update them with residual connections.

For $N$ feature maps $\mathbf{F} = \{f_i, i=1,...,N\}$, FPM will also generate $N$ polished features maps  $\mathbf{F}^p = \{f_i^p, i=1,...,N\}$ with the same size.
As is shown in Fig.~\ref{pipeline},
FPM consists of $N$ parallel FPM blocks, each of which corresponds to a separate feature map and is denoted as FPM-$k$.
Specifically, a series of short connections~\cite{hou2017deeply} from deeper side to shallower side are adopted.
As a result, higher level features with global information are injected directly to lower ones to help better discover the salient regions.
Taking the FPM1-3 in Fig.~\ref{pipeline} as an example, all features of \textit{Res-3}, \textit{Res-4}, \textit{Res-5} are utilized through short connections to update the features of \textit{Res-3}.
FPM also takes advantage of residual connections~\cite{he2016deep} so that it can update the features and gradually filter out the cluttered information.
This is illustrated by the connection surrounding each FPM block in Fig.~\ref{pipeline}.

\begin{figure}
	\centering
	\includegraphics[width=0.98\linewidth,height=0.781\linewidth]{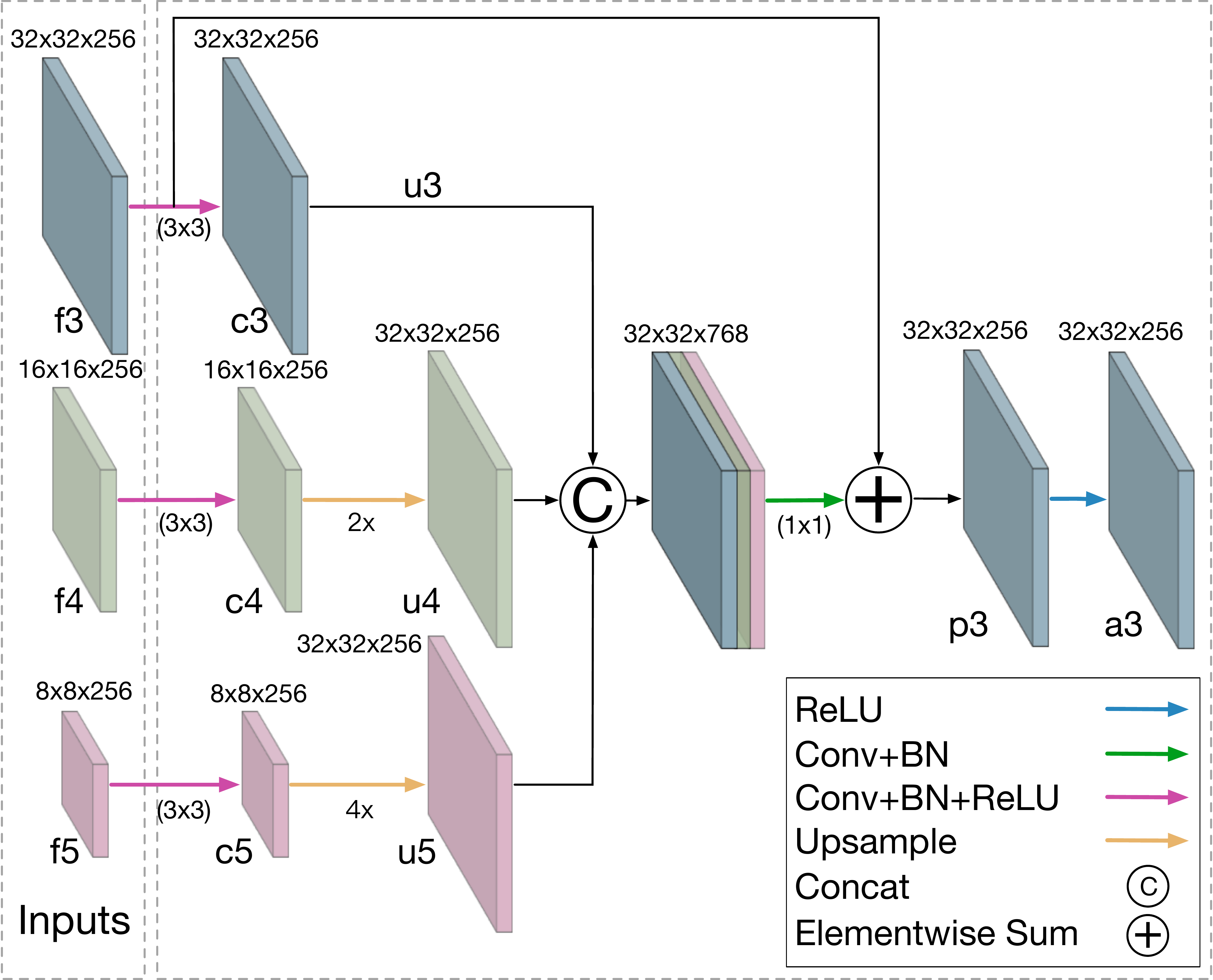}\\
	\caption{
		\small Illustration of the detail implementation of a FPM block with a residual connection, which is formally formulated by Eq. \ref{equa_fpm}. This is an example of $N=5, k=3$, i.e. FPM1-3 and FPM2-3 in Fig.~\ref{pipeline}.}
	\label{fig:fpm}
\end{figure}

The implementation of FPM-$k$ block is formally formulated as Eq. \ref{equa_fpm}:
\begin{equation}
\label{equa_fpm}
\begin{split}
	&c_j =\mathrm{ReLU}(\mathrm{BN}(\mathrm{Conv}(f_j))) \\
	& \quad \quad for \quad j=k,k+1,..,N \\
	&u_j = \left\{
		\begin{array}{lr}
			\mathrm{Upsample}(c_j) \quad  j>k \\
			c_j \quad \quad \quad \quad \quad \quad j=k
		\end{array}
		\right. \\
&p_k =\mathrm{BN}(\mathrm{Conv}(\mathrm{Concat}(u_k,u_{k+1}..., u_N))) \\
&f_k^p = \mathrm{ReLU}(p_k + f_k)
\end{split}
\end{equation}
It takes in $N-k+1$ feature maps, i.e. $\{f_j, j = k,...,N\}$.
For feature map $f_j$, we first apply a convolutional layer with a 3x3 kernel followed by a batch normalization and a ReLU non-linearity to capture context knowledge,
and interpolate it bilinearly to the size of $f_k$.
These features are then combined with a concatenation along channels and fused by a 1x1 convolutional layer to reduce the dimension, obtaining $p_k$.
Finally, $p_k$ is used as a residual function to update the original feature map $f_k$ to compute the $f_k^p$ with element-wise addition.
An example of this procedure with $k=3$ is illustrated in Fig.~\ref{fig:fpm}.

\subsection{Fusion Module}
\label{approach:fm}
We use the Fusion Module~(FM) to finally integrate the multi-level features and detect salient objects.
As result of our refined features, the FM can be quite simple.
As is illustrated in Fig.~\ref{pipeline}, the multi-level features from TM2 are first concatenated and then fed into two successive convolutional layers with 3x3 kernels.
At last, a 1x1 convolutional layer followed by a sigmoid function is applied to obtain the final saliency map.



\subsection{Implementation Details}
\label{approach:details}
We use the cross-entropy loss between the final predicted saliency map and ground truth to train our model end-to-end.
Following previous works~\cite{hou2017deeply,liu2018picanet,zhuge2019def}, side outputs are also employed to calculate auxiliary losses.
In detail, 1x1 convolutional layers are performed on the multi-level feature maps before the Fusion Module to obtain a series of intermediate results.
The total loss is as follows:
\begin{equation}
L_{total}=\mathcal{L}(s, g)+0.5\mathcal{L}(s_1, g)+0.3 \sum_{i=2}^N\mathcal{L}(s_i, g)
\end{equation}
where $s$ is the final result of our model, $s_i$ denotes the $i$-th intermediate result, and $g$ represents the ground truth. The weights are set empirically to bias towards the final result.

We implement our method with Pytorch~\cite{paszke2017automatic} framework.
The last average pooling layer and fully connected layer of the pre-trained ResNet-101~\cite{he2016deep} are removed.
We initialize the layers of backbone with the weights pre-trained on ImageNet classification task and randomly initialize the rest layers.
We follow source code of PiCA~\cite{liu2018picanet} given by author and FQN~\cite{Li_2019_CVPR} and freeze the BatchNorm statistics of the backbone.



\begin{table*}[hbt]
	\centering
		\caption{
			\small
        Quantitative comparisons with different methods on 5 datasets with MAE (smaller is better), max/mean F-measure score (larger is better) and S-measure (larger is better).
        The best three results are shown in {\color{red}red}, {\color{blue}blue} and {\color{green}green}.
        The results of our method with $T=2$ based on both ResNet101~\cite{he2016deep} and VGG16~\cite{simonyan2014very} are reported.}
    \tiny
    \label{table:metrics_comparision}
    \renewcommand\tabcolsep{3pt}
	\begin{tabular}{l|cccc|cccc|cccc|cccc|cccc}
		\hline
		\multirow{2}*{Method}&\multicolumn{4}{c|}{ECSSD}&\multicolumn{4}{c|}{HKU-IS}&\multicolumn{4}{c|}{DUT-O}&\multicolumn{4}{c|}{DUTS-TE}&\multicolumn{4}{c}{PASCAL-S}\cr
		&MAE&max\ F&mean\ F&S  &MAE&max\ F&mean\ F&S  &MAE&max\ F&mean\ F&S &MAE&max\ F&mean\ F&S &MAE&max\ F&mean\ F&S\cr
		\hline\hline
		\multicolumn{21}{c}{VGG~\cite{simonyan2014very}}\\
		\hline
		RFCN~\cite{wang2016saliency}    &0.107&0.890&0.811&0.852     &0.079&0.892&0.805&0.859   &0.111&0.742&0.656&0.764   &0.091&0.784&0.728&0.794    &0.118&0.837&0.785&0.804\\
		DHS~\cite{liu2016dhsnet}        &0.059&0.907&0.885&0.883     &0.053&0.890&0.867&0.869   &-    &-    &-    &-       &0.067&0.807&0.777&0.817    &0.094&0.842&0.829&0.802\\
		RAS~\cite{chen2018reverse}      &0.056&0.921&0.900&0.893     &0.045&0.913&0.887&0.887   &0.062&0.786&0.762&0.813   &0.060&0.831&0.803&0.838    &0.104&0.837&0.829&0.785\\
		Amulet~\cite{zhang2017amulet}   &0.059&0.915&0.882&0.893     &0.052&0.895&0.856&0.883   &0.098&0.742&0.693&0.780   &0.085&0.778&0.731&0.804    &0.098&0.837&0.838&0.822\\
		DSS~\cite{hou2017deeply}        &0.052&0.916&0.911&0.882     &0.041&0.910&0.904&0.879   &0.066&0.771&0.764&0.787   &0.057&0.825&0.814&0.824    &0.096&0.852&0.849&0.791\\
        PiCA~\cite{liu2018picanet}      &0.047&0.931&0.899&0.913     &0.042&0.921&0.883&0.905   &0.068&0.794&0.756&0.820   &0.054&0.851&0.809&0.858    &0.088&0.880&0.854&0.842\\
        BMPM~\cite{zhang2018bi}         &0.045&0.929&0.900&0.911     &0.039&0.921&0.888&0.905   &0.064&0.774&0.744&0.808   &0.049&0.851&0.814&0.861    &0.074&0.862&0.855&0.834\\
        AFN~\cite{Feng_2019_CVPR}       &0.042&0.935&0.915&0.914     &0.036&0.923&0.899&0.905   &\color{green}\textbf{0.057}&0.797&0.776&0.826   &0.046&0.862&0.834&0.866    &0.076&0.879&0.866&0.841\\
		CPD~\cite{wu2019cascaded}       &0.040&0.936&0.923&0.910     &0.033&0.924&0.903&0.904   &\color{green}\textbf{0.057}&0.794&0.780&0.817   &0.043&0.864&0.846&0.866    &0.074&0.877&0.868&0.832\\
        MLMS~\cite{wu2019mutual}        &0.044&0.928&0.900&0.911     &0.039&0.921&0.888&0.906   &0.063&0.774&0.745&0.808   &0.048&0.846&0.815&0.861    &0.079&0.877&0.857&0.836\\
        ICTBI~\cite{wang2019iterative}  &0.041&0.921&-&-             &0.040&0.919&-&-           &0.060&0.770&-    &-       &0.050&0.830&-&-            &0.073&0.840&-&-\\
        ours                            &0.040&0.938&0.915&\color{green}\textbf{0.916}
		                                &0.035&\color{blue}\textbf{0.928}&0.902&\color{blue}\textbf{0.909}
										&0.063&0.777&0.753&0.805
										&\color{blue}\textbf{0.042}&\color{blue}\textbf{0.868}&0.836&0.864
	 	                                &\color{green}\textbf{0.071}&\color{blue}\textbf{0.891}&\color{blue}\textbf{0.866}&0.834\\
		\hline
		\multicolumn{21}{c}{ResNet~\cite{he2016deep}}\\
		\hline
		SRM~\cite{wang2017stagewise}    &0.054&0.917&0.896&0.895     &0.046&0.906&0.881&0.886   &0.069&0.769&0.744&0.797   &0.059&0.827&0.796&0.836    &0.085&0.847&0.847&0.830\\
		PiCA~\cite{liu2018picanet}      &0.047&0.935&0.901&\color{blue}\textbf{0.918}    &0.043&0.919&0.880&0.904     &0.065&\color{green}\textbf{0.803}&0.762&\color{green}\textbf{0.829}   &0.051&0.860&0.816&\color{green}\textbf{0.868}    &0.077&\color{green}\textbf{0.881}&0.851&\color{blue}\textbf{0.845}\\
		DGRL~\cite{wang2018detect}      &0.041&0.922&0.912&0.902     &0.036&0.910&0.899&0.894   &0.062&0.774&0.765&0.805   &0.050&0.829&0.820&0.842    &0.072&0.872&0.854&0.831\\
        CAPS~\cite{zhang2019capsal}     &-    &-    &-    &-         &0.057&0.882&0.865&0.852   &-&-&-&-                   &0.060&0.821&0.802&0.819    &0.078&0.866&0.860&0.826\\
        BAS~\cite{qin2019basnet}        &{\color{green}\textbf{0.037}}&{\color{blue}\textbf{0.942}}&\color{red}\textbf{0.927}&\color{green}\textbf{0.916}     &\color{blue}\textbf{0.032}&\color{blue}\textbf{0.928}&\color{blue}\textbf{0.911}&\color{green}\textbf{0.908}   &\color{blue}\textbf{0.056}&\color{blue}\textbf{0.805}&\color{blue}\textbf{0.790}&\color{blue}\textbf{0.835}   &0.047&0.855&\color{green}\textbf{0.842}&0.865    &0.084&0.872&0.861&0.824\\
        ICTBI~\cite{wang2019iterative}  &0.040&0.926&-&-             &0.038&0.920&-&-           &0.059&0.780&-&-           &0.048&0.836&-&-            &0.072&0.848&-&-\\
        CPD~\cite{wu2019cascaded}       &{\color{green}\textbf{0.037}}&{\color{green}\textbf{0.939}}&\color{green}\textbf{0.924}&\color{blue}\textbf{0.918}     &0.034&\color{green}\textbf{0.925}&0.904&0.905   &\color{blue}\textbf{0.056}&0.797&\color{green}\textbf{0.780}&0.824   &\color{green}\textbf{0.043}&\color{green}\textbf{0.865}&\color{blue}\textbf{0.844}&\color{blue}\textbf{0.869}    &0.078&\color{green}\textbf{0.876}&\color{green}\textbf{0.865}&\color{green}\textbf{0.835}\\
        DEF~\cite{zhuge2019def}         &\color{blue}\textbf{0.036}&-&0.915&-   &\color{green}\textbf{0.033}&-&\color{green}\textbf{0.907}&-   &0.062&-&0.769&-      &0.045&-&0.821&-   &\color{blue}\textbf{0.070}&-&0.826&-\\
		ours                            &{\color{red}\textbf{0.033}}&{\color{red}\textbf{0.949}}&{\color{blue}\textbf{0.926}}&{\color{red}\textbf{0.932}}
                                        &{\color{red}\textbf{0.030}}&{\color{red}\textbf{0.939}}&{\color{red}\textbf{0.912}}&{\color{red}\textbf{0.921}}
										&{\color{red}\textbf{0.053}}&{\color{red}\textbf{0.820}}&{\color{red}\textbf{0.794}}&{\color{red}\textbf{0.842}}
										&{\color{red}\textbf{0.037}}&{\color{red}\textbf{0.888}}&{\color{red}\textbf{0.858}}&{\color{red}\textbf{0.887}}
										&{\color{red}\textbf{0.068}}&{\color{red}\textbf{0.892}}&{\color{red}\textbf{0.873}}&{\color{red}\textbf{0.851}}\\
		\hline
    \end{tabular}

\end{table*}

\begin{figure*}[hbt]
	\centering
	\setlength{\tabcolsep}{-1mm}
	\begin{tabular}{ccccc}
			\includegraphics[width=3.7cm]{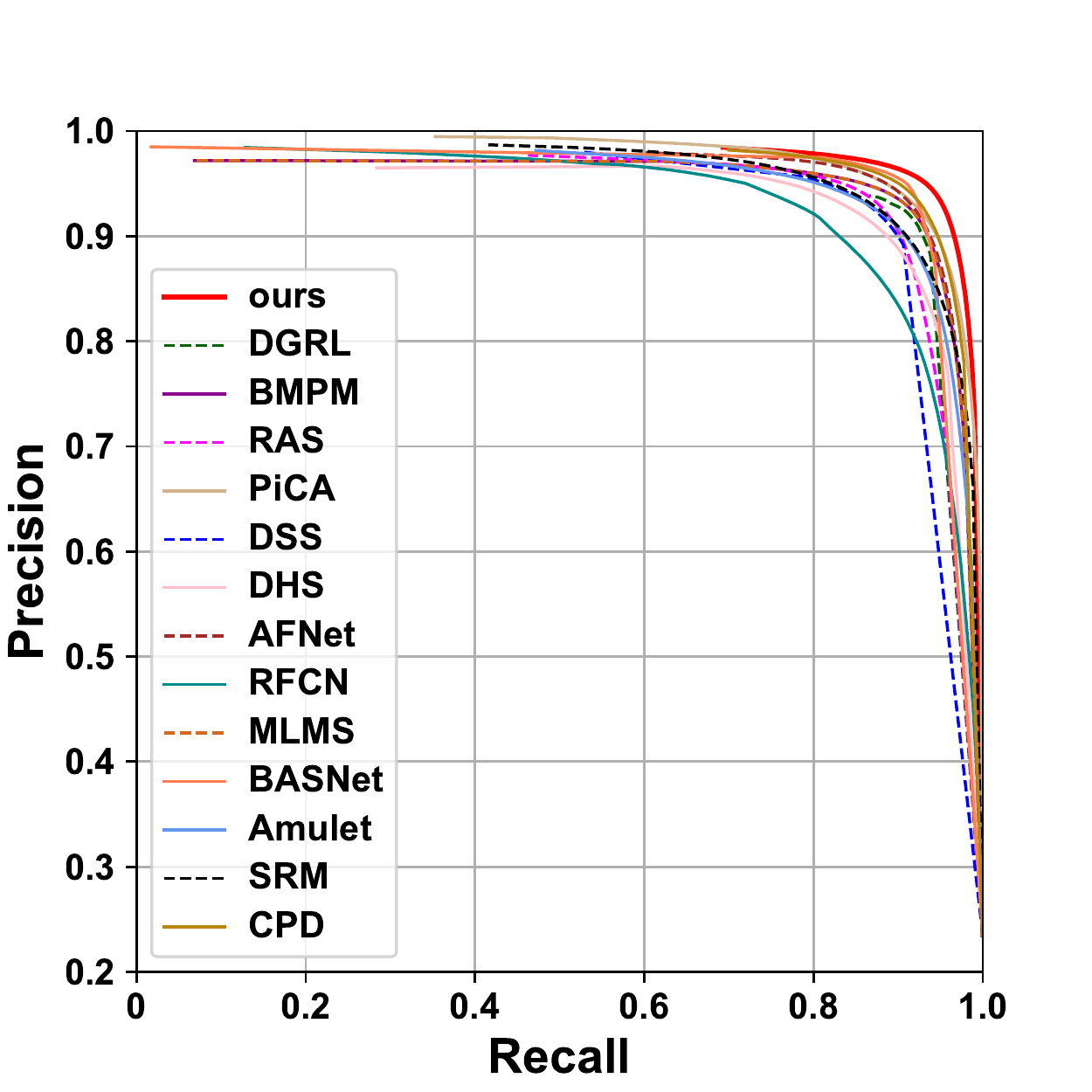}
			&\includegraphics[width=3.7cm]{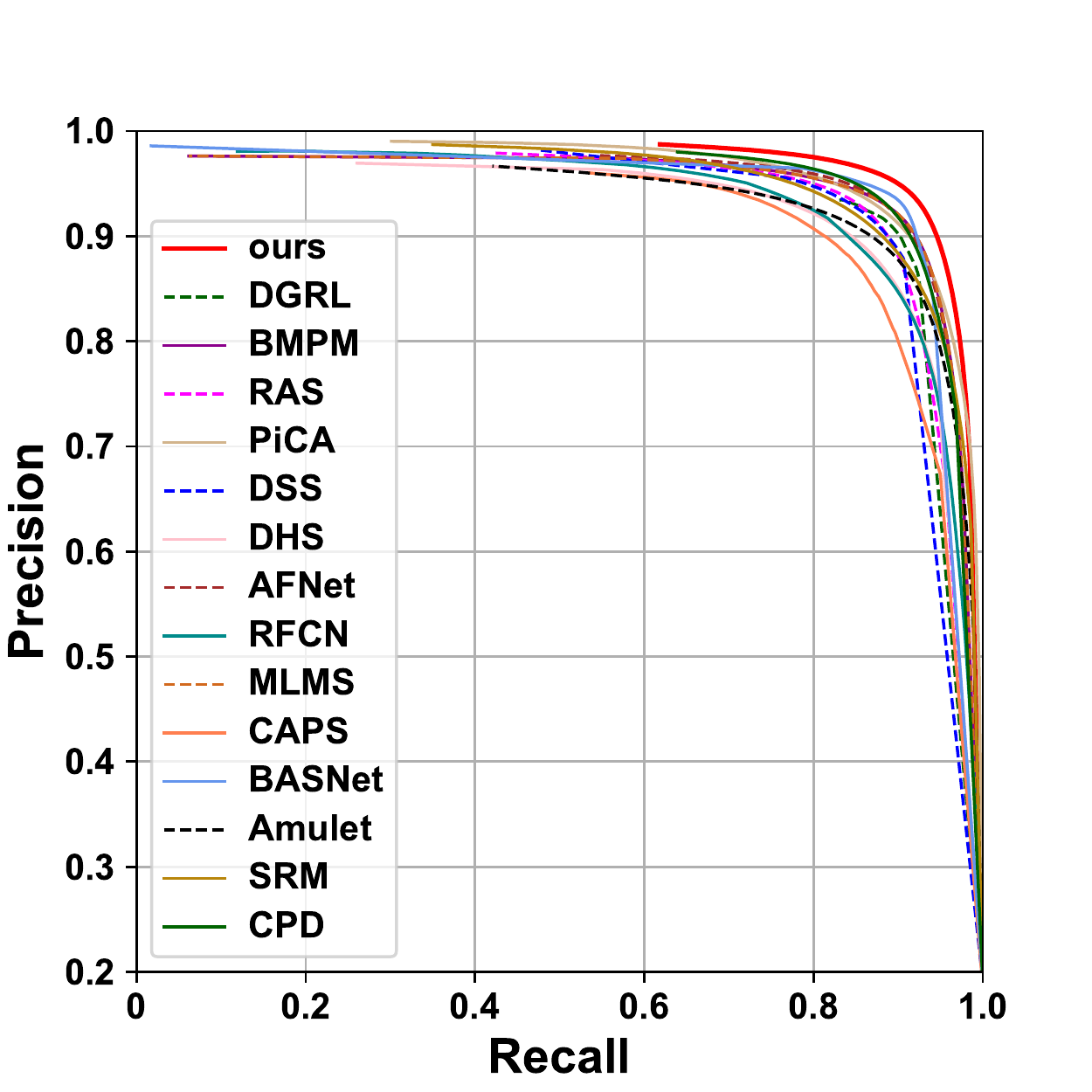}
			&\includegraphics[width=3.7cm]{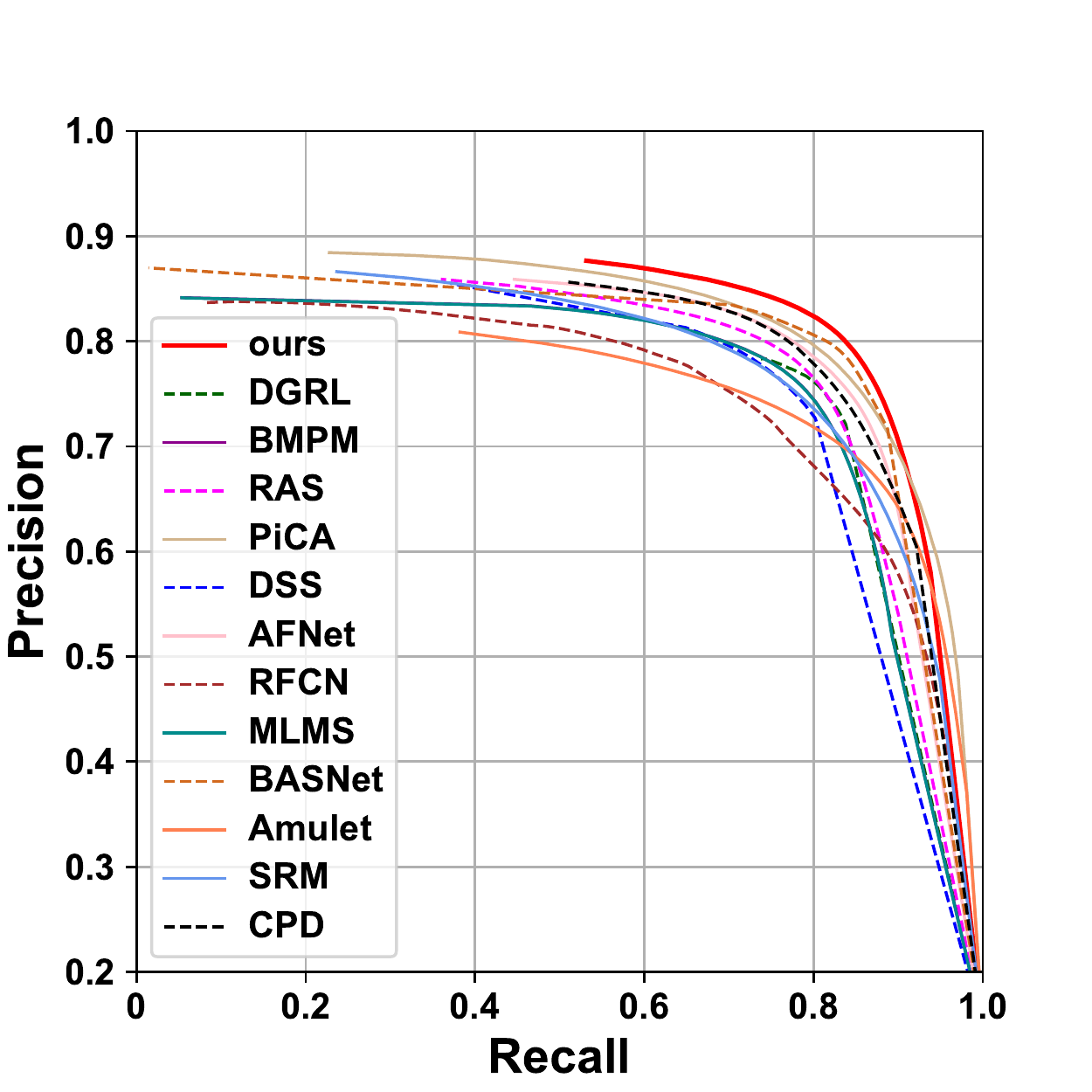}
			&\includegraphics[width=3.7cm]{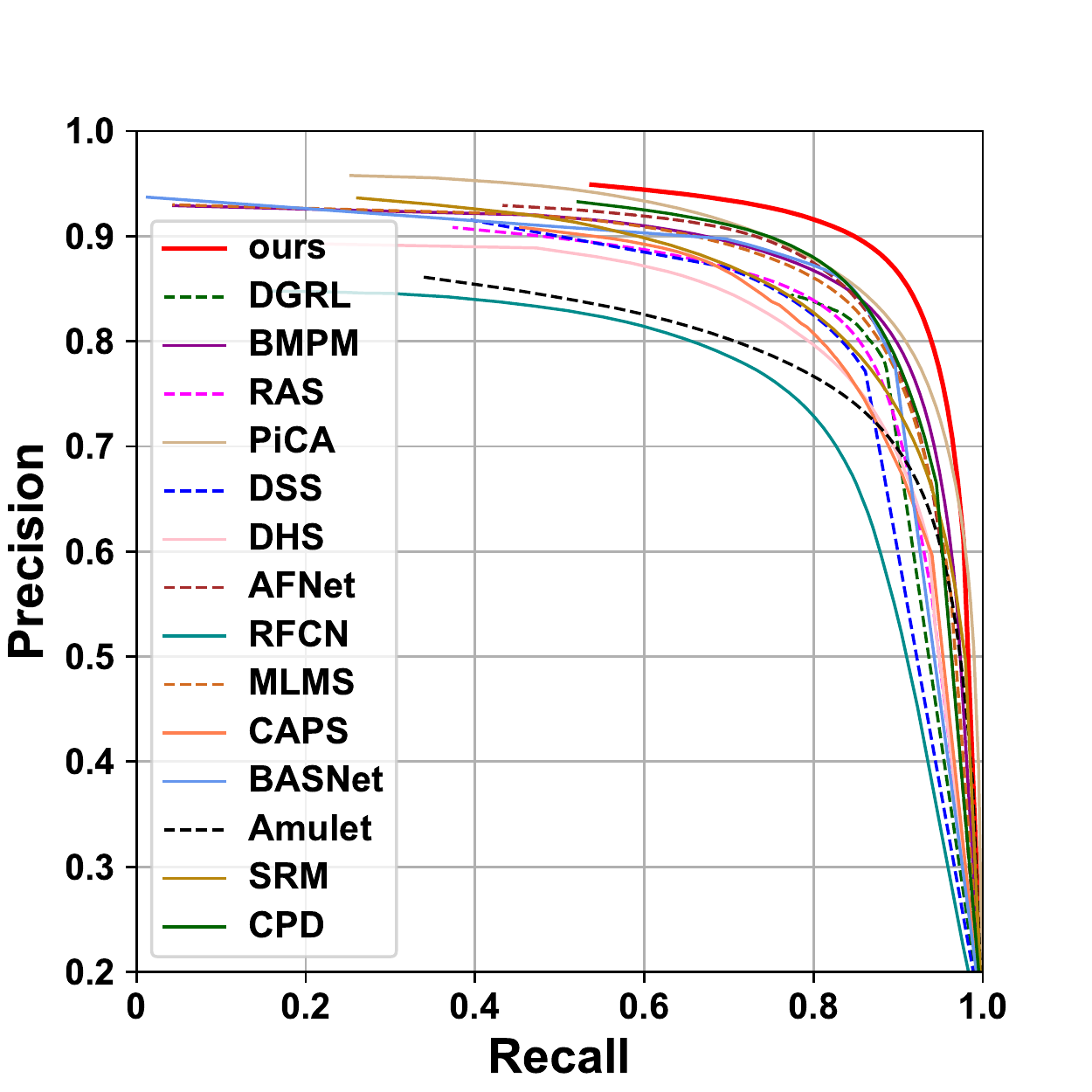}
			&\includegraphics[width=3.7cm]{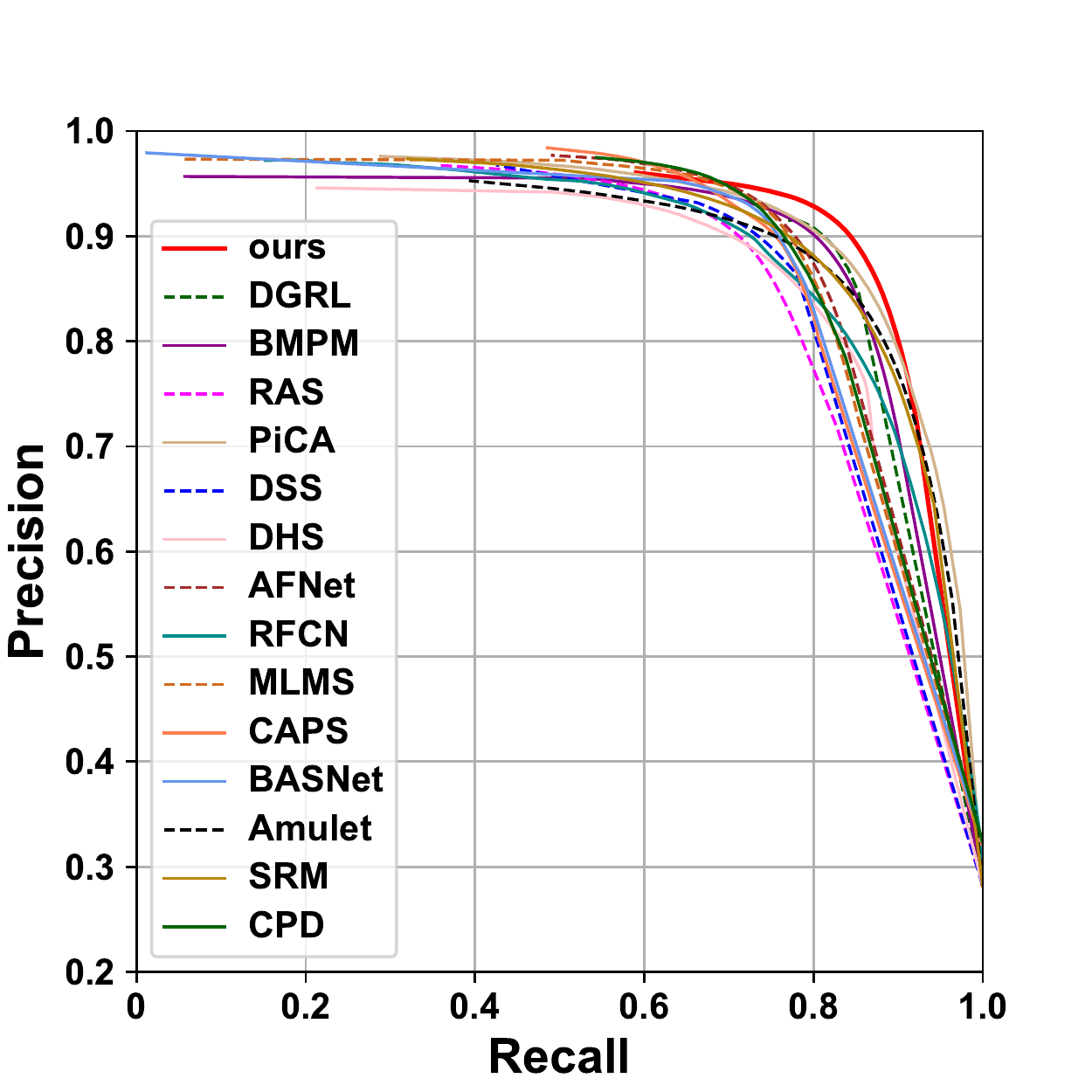}\\
			ECSSD & HKU-IS & DUT-O & DUTS-TE & PASCAL-S
			
	\end{tabular}
	\caption{
		\small PR curves with different thresholds of our method and other state-of-art methods on five benchmark datasets.}
	\label{fig:PR_comparision}
\end{figure*}

\section{Experiments}

\subsection{Datasets and metrics}
We conduct experiments on five well-known benchmark datasets:  ECSSD, HKU-IS, PASCAL-S, DUT-OMRON and DUTS.
{\bf ECSSD}~\cite{yan2013hierarchical} consists of 1,000 images. This dataset contains salient objects
with complex structures in multiple scales. 
{\bf HKU-IS}~\cite{li2015visual} consists of 4,447 images and most images are chosen to contain mutliple disconnected salient objects.
{\bf PASCAL-S}~\cite{li2014secrets} includes 850 natural images. These images are selected from PASCAL VOC 2010 segmentation challenge and are pixel-wise annotated.
{\bf DUT-O}~\cite{yang2013saliency} is a challenging dataset in that each image contains one or more salient objects with fairly complex scenes. This dataset has 5,168 high-quality images. 
{\bf DUTS}~\cite{wang2017learning} is a large scale dataset which consists of 15,572 images, which are selected from ImageNet DET~\cite{deng2009imagenet} and SUN~\cite{xiao2010sun} dataset. It has been split into two parts: 10,553 for training and 5,019 for testing.
We evaluate the performance of different salient object detection algorithms through 4 main metrics,
including the precision-recall curves (PR curves), F-measure, mean absolute error(MAE), S-measure~\cite{FanStructMeasureICCV17}.
By binarizing the predicted saliency map with thresholds in [0,255], a sequence of precision and recall pairs are calculated for each image of the dataset. 
The PR curve is plotted using the average precision and recall of the dataset at different thresholds.
F-measure is calculated as a weighted combination of Precision and Recall with the formulation as follows:
\begin{equation}
    F_\beta = \frac{(1+\beta^2)Precision + Recall}{\beta^2Precision + Recall}
\end{equation}
where $\beta^2$ is usually set to $0.3$ to emphasize Precision more than Recall as suggested in \cite{yang2013saliency}.

\subsection{Training and Testing}
Following the conventional practice~\cite{liu2018picanet,zhang2018bi,zhang2018bi}, our proposed model is trained on the training set of DUTS dataset.
We also perform a data augmentation similar to \cite{liu2018picanet} during training to mitigate the over-fitting problem.
Specifically, the image is first resized to 300x300 and then a 256x256 image patch is randomly cropped from it.
Random horizontal flipping is also applied.
We use Adam optimizer to train our model without evaluation until the training loss convergences.
The initial learning rate is set to 1e-4 and the overall training procedure takes about 16000 iterations.
For testing, the images are scaled to 256x256 to feed into the network and then the predicted saliency maps are bilinearly interpolated to the size of the original image.

\begin{figure*}
	\centering
	\setlength{\tabcolsep}{0.2mm}
	\renewcommand{\arraystretch}{0.7} 
	\begin{tabular}{ccccccccccccccccc}
		\includegraphics[width=1.4cm]{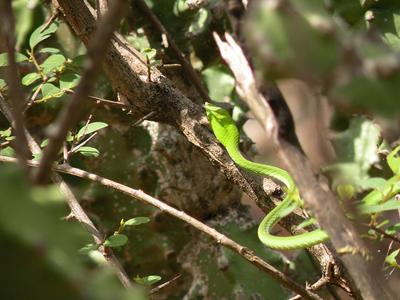}
		&\includegraphics[width=1.4cm]{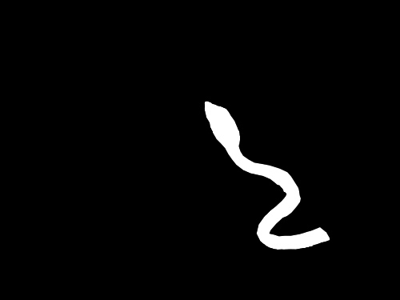}
        &\includegraphics[width=1.4cm]{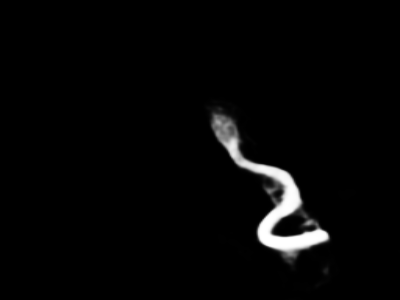}
        &\includegraphics[width=1.4cm]{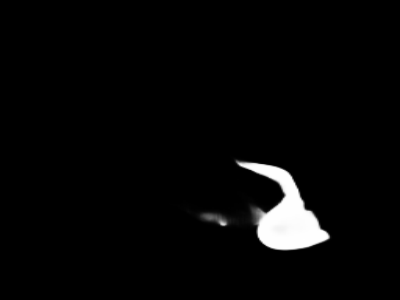}
        &\includegraphics[width=1.4cm]{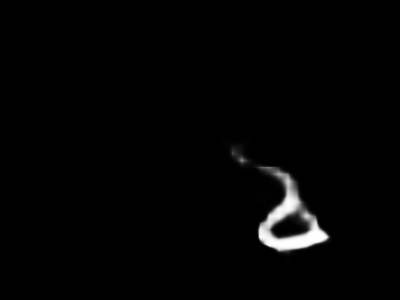}
		&\includegraphics[width=1.4cm]{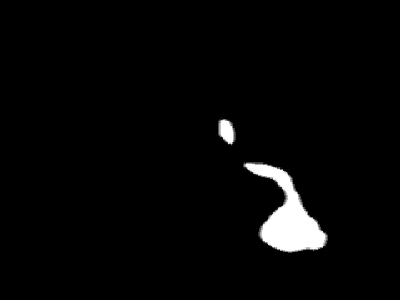}
		&\includegraphics[width=1.4cm]{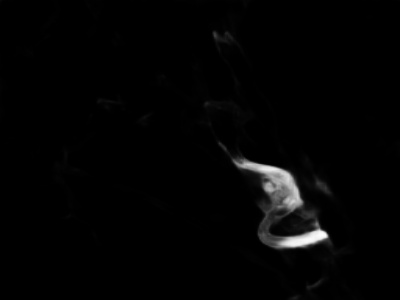}
		&\includegraphics[width=1.4cm]{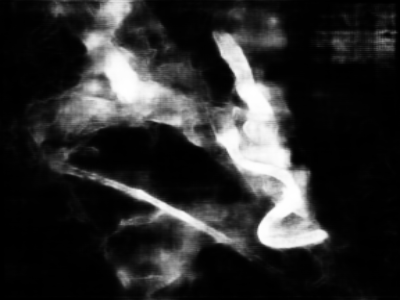}
		&\includegraphics[width=1.4cm]{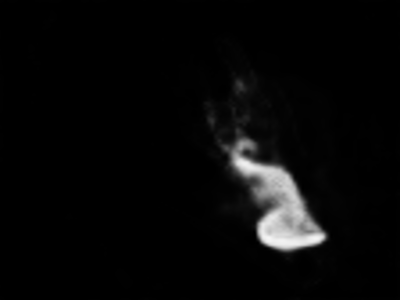}
		&\includegraphics[width=1.4cm]{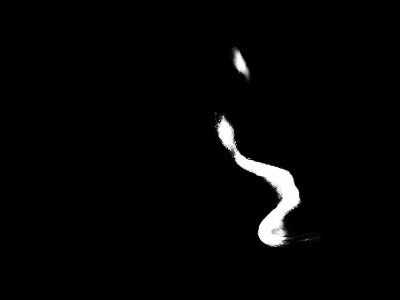}
		&\includegraphics[width=1.4cm]{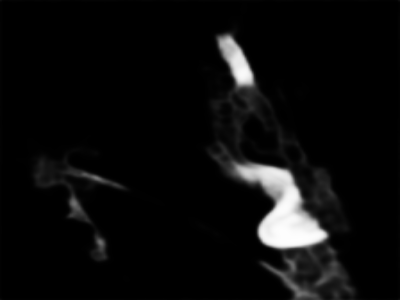}
		&\includegraphics[width=1.4cm]{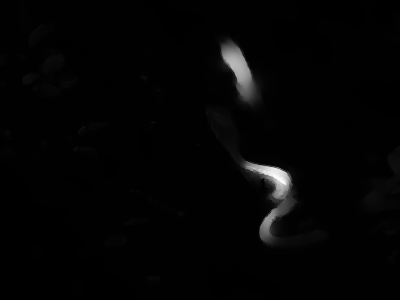}\\
		
		\includegraphics[width=1.4cm]{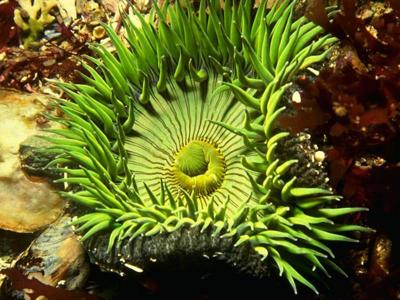}
		&\includegraphics[width=1.4cm]{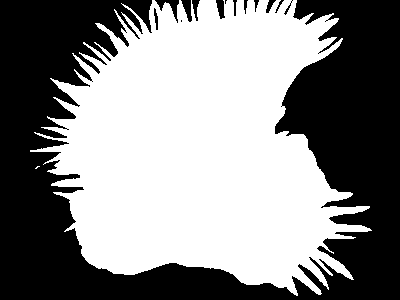}
        &\includegraphics[width=1.4cm]{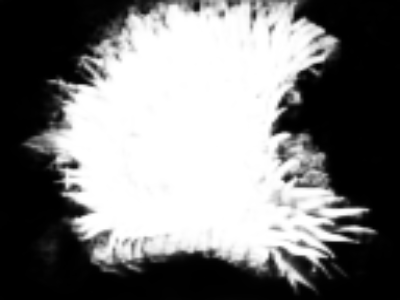}
        &\includegraphics[width=1.4cm]{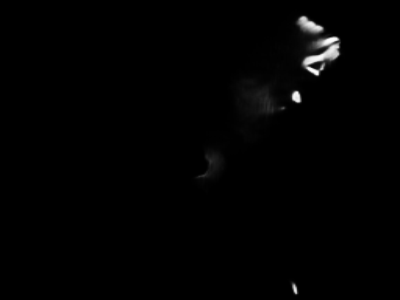}
        &\includegraphics[width=1.4cm]{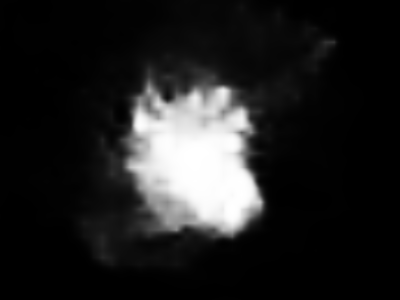}
		&\includegraphics[width=1.4cm]{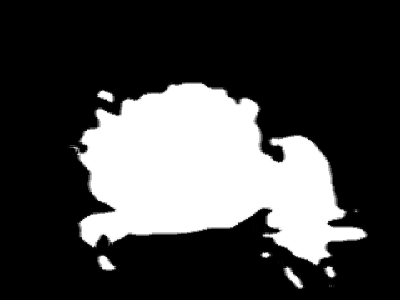}
		&\includegraphics[width=1.4cm]{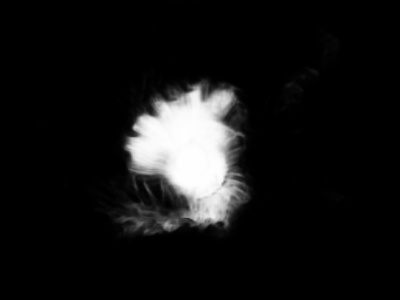}
		&\includegraphics[width=1.4cm]{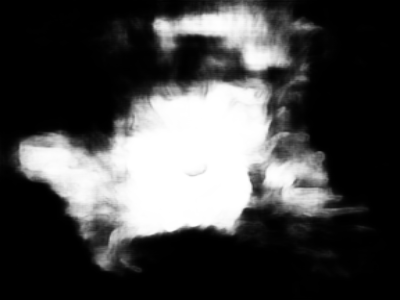}
		&\includegraphics[width=1.4cm]{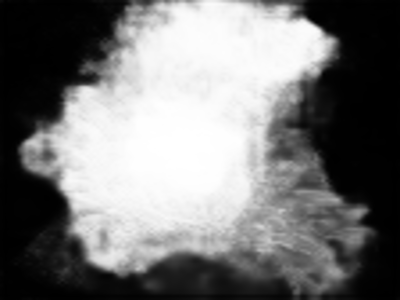}
		&\includegraphics[width=1.4cm]{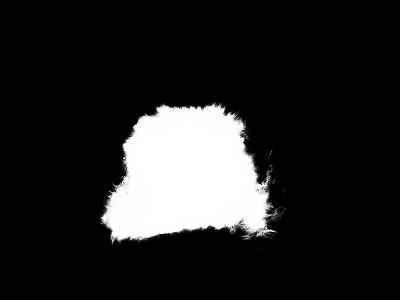}
		&\includegraphics[width=1.4cm]{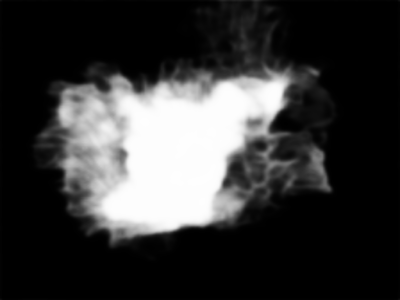}
		&\includegraphics[width=1.4cm]{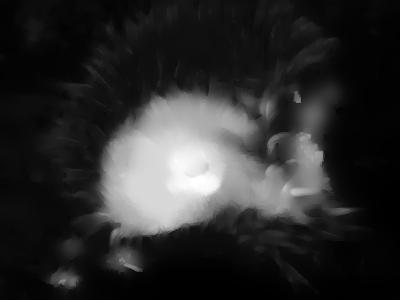}\\

		\includegraphics[width=1.4cm]{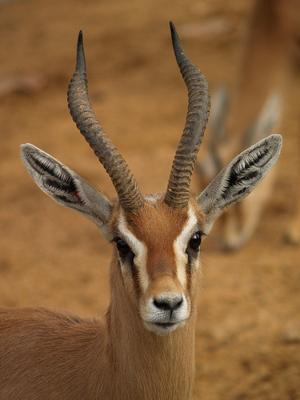}
		&\includegraphics[width=1.4cm]{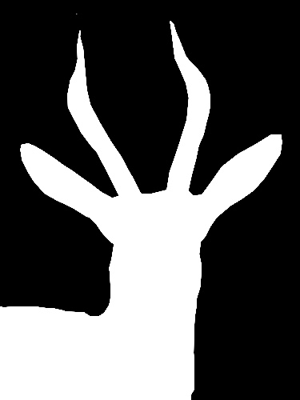}
        &\includegraphics[width=1.4cm]{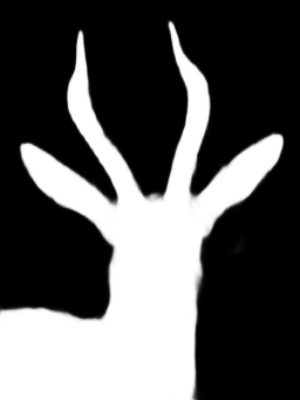}
        &\includegraphics[width=1.4cm]{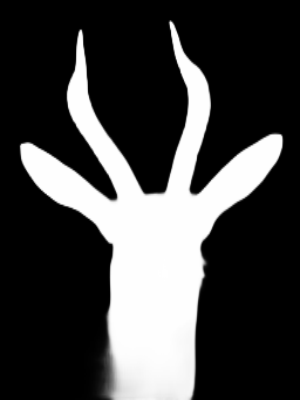}
        &\includegraphics[width=1.4cm]{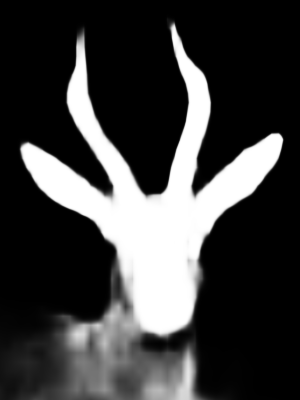}
		&\includegraphics[width=1.4cm]{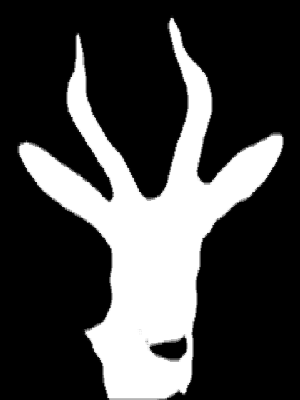}
		&\includegraphics[width=1.4cm]{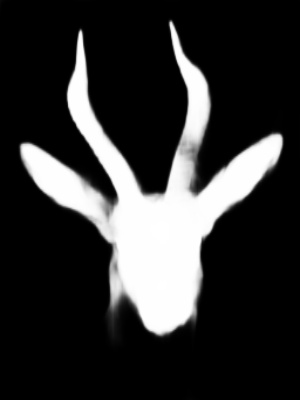}
        &\includegraphics[width=1.4cm]{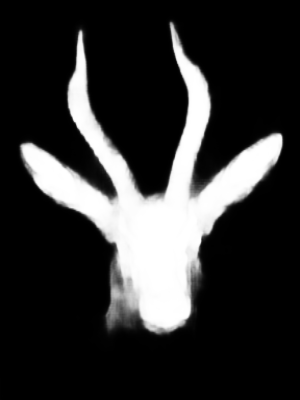} 
		&\includegraphics[width=1.4cm]{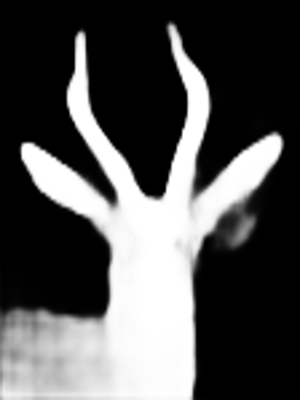}
		&\includegraphics[width=1.4cm]{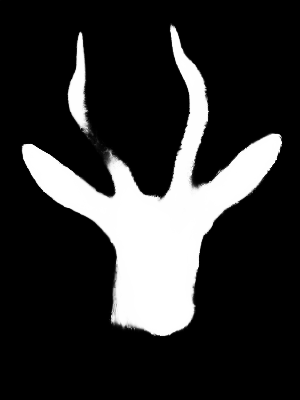}
		&\includegraphics[width=1.4cm]{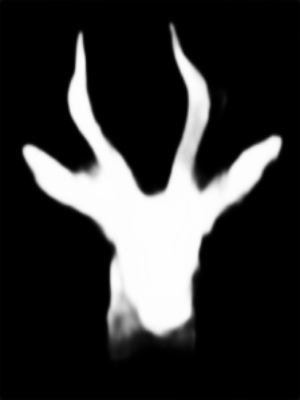}
		&\includegraphics[width=1.4cm]{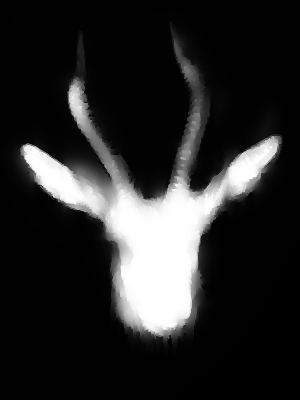}\\

		\includegraphics[width=1.4cm]{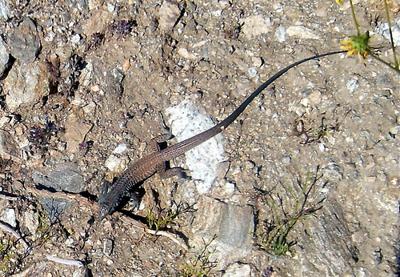}
		&\includegraphics[width=1.4cm]{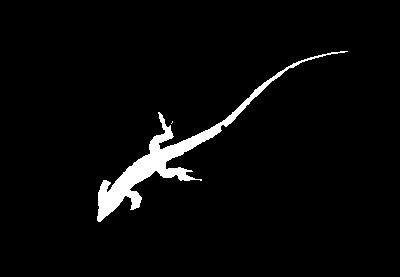}
        &\includegraphics[width=1.4cm]{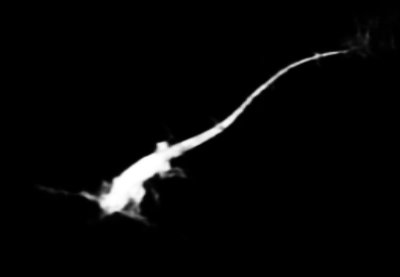}
        &\includegraphics[width=1.4cm]{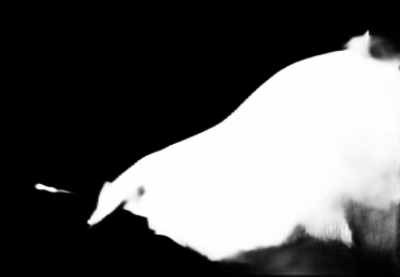}
        &\includegraphics[width=1.4cm]{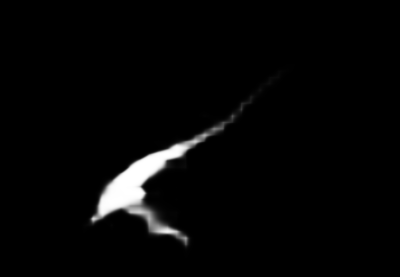}
		&\includegraphics[width=1.4cm]{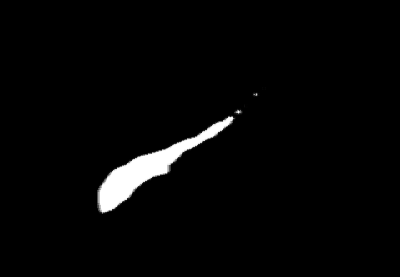}
		&\includegraphics[width=1.4cm]{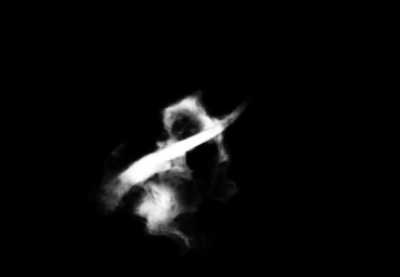}
        &\includegraphics[width=1.4cm]{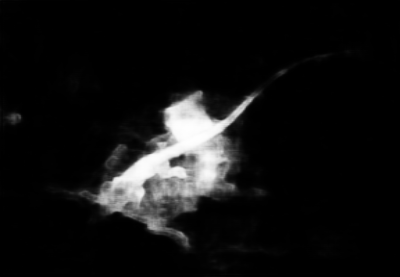}
		&\includegraphics[width=1.4cm]{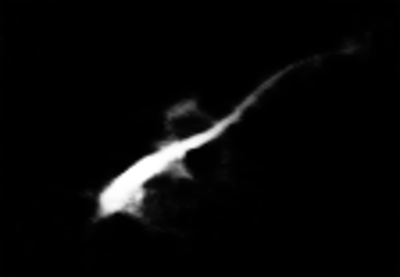}
		&\includegraphics[width=1.4cm]{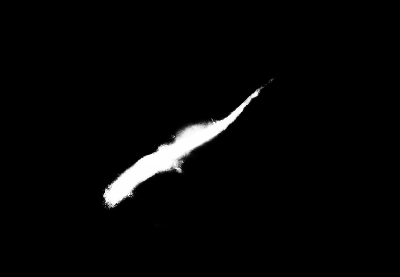}
		&\includegraphics[width=1.4cm]{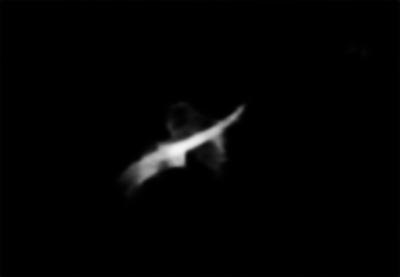}
		&\includegraphics[width=1.4cm]{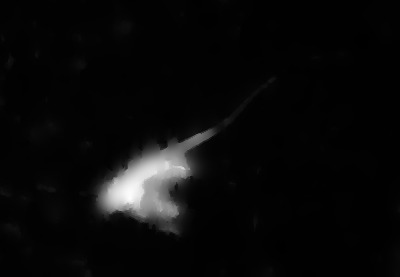}\\
		
		
		\includegraphics[width=1.4cm]{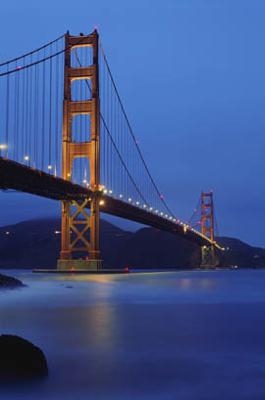}
		&\includegraphics[width=1.4cm]{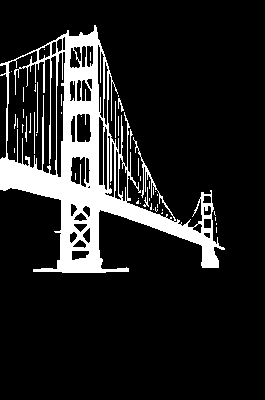}
        &\includegraphics[width=1.4cm]{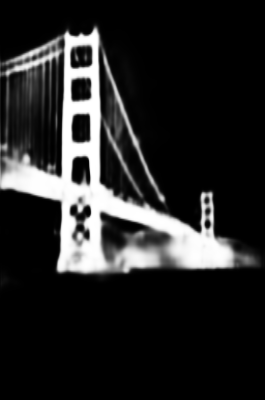}
        &\includegraphics[width=1.4cm]{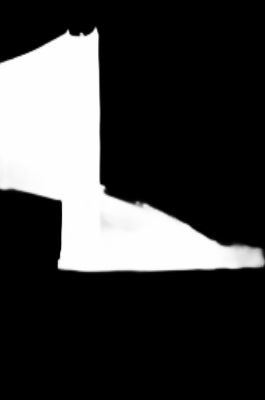}
        &\includegraphics[width=1.4cm]{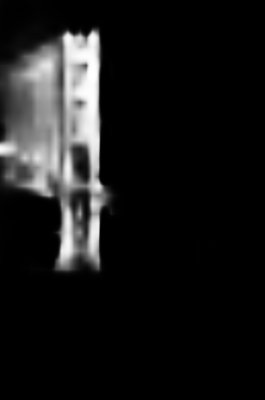}
		&\includegraphics[width=1.4cm]{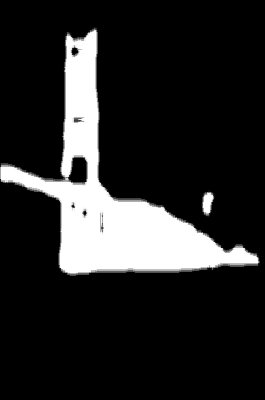}
		&\includegraphics[width=1.4cm]{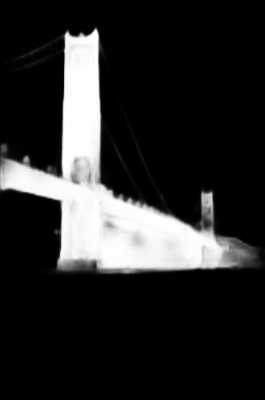}
		&\includegraphics[width=1.4cm]{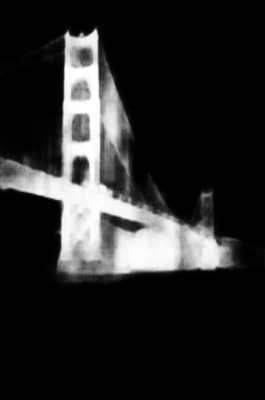}
		&\includegraphics[width=1.4cm]{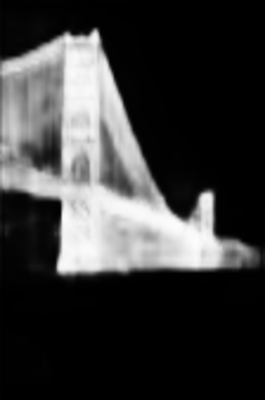}
		&\includegraphics[width=1.4cm]{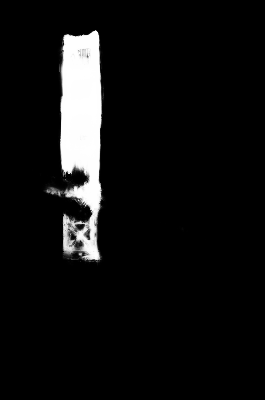}
		&\includegraphics[width=1.4cm]{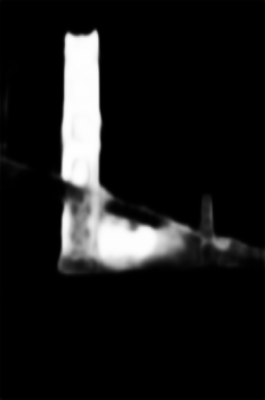}
		&\includegraphics[width=1.4cm]{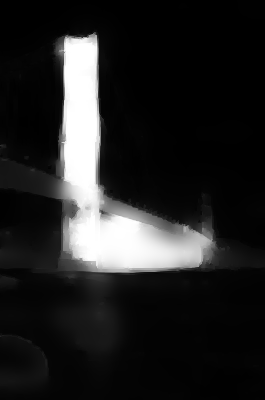}\\
		
		\includegraphics[width=1.4cm]{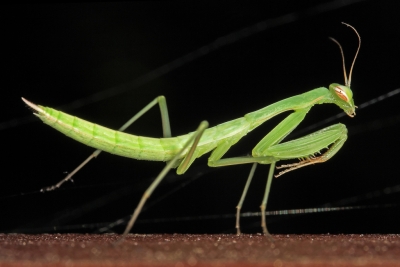}
		&\includegraphics[width=1.4cm]{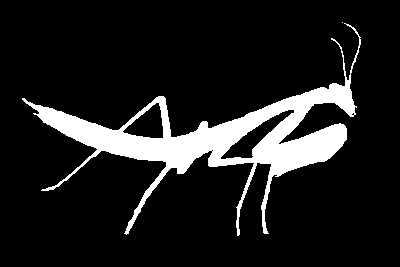}
        &\includegraphics[width=1.4cm]{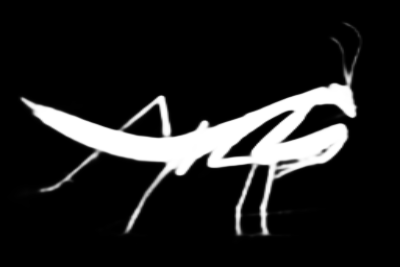}
        &\includegraphics[width=1.4cm]{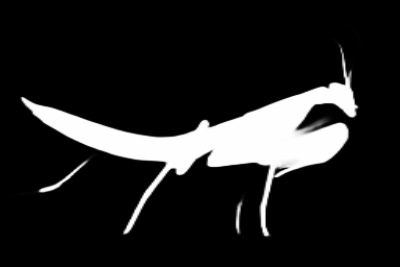}
        &\includegraphics[width=1.4cm]{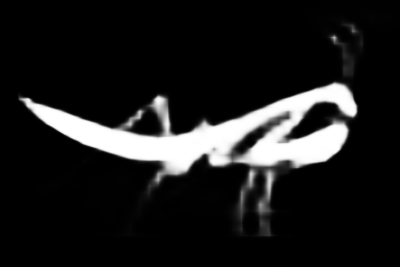}
		&\includegraphics[width=1.4cm]{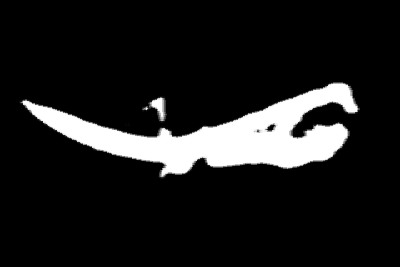}
		&\includegraphics[width=1.4cm]{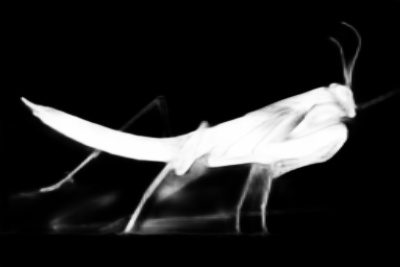}
		&\includegraphics[width=1.4cm]{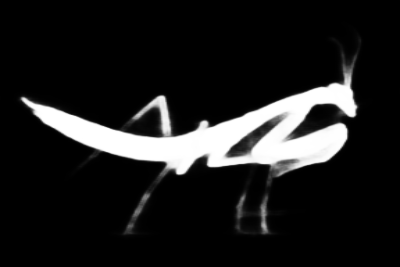}
		&\includegraphics[width=1.4cm]{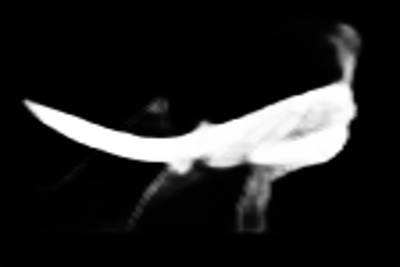}
		&\includegraphics[width=1.4cm]{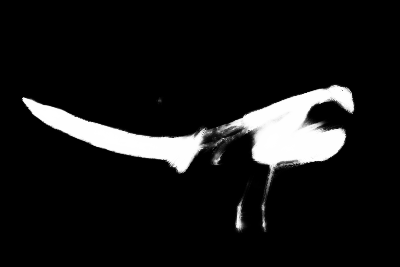}
		&\includegraphics[width=1.4cm]{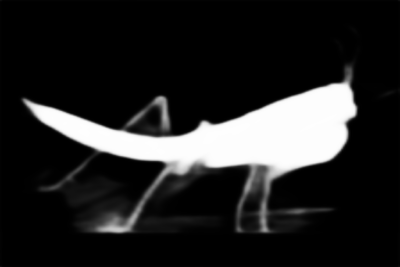}
		&\includegraphics[width=1.4cm]{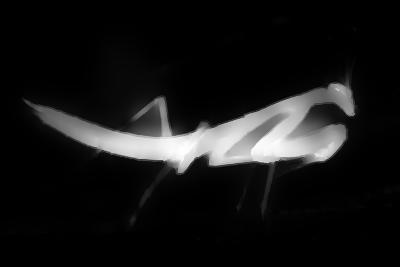}\\
		
		
		Input & GT & ours & BAS & CPD & DGRL & BMPM & Amulet & PiCA-R & DSS & DHS & RFCN \\
		
	\end{tabular}
	\caption{\small Visual comparison with different methods in various scenarios.}
	\label{fig:visual_comparision}
\end{figure*}

\subsection{Comparison with the state-of-the-art}
We compare our proposed model with 16 state-of-the-art methods.
For fair comparison, the metrics of these 16 methods are obtained from a public leaderboard~\cite{sal_eval_toolbox} or their original papers, and we evaluate our method in the same way as~\cite{sal_eval_toolbox}.
We report the results of our model with ResNet-101~\cite{he2016deep} as backbone and two FPMs~(i.e. $T=2$) if not otherwise mentioned.
The saliency maps for visual comparisons are provided by the authors.

\textbf{Quantitative Evaluation.}
The quantitative performances of all methods can be found in Table~\ref{table:metrics_comparision} and Fig.~\ref{fig:PR_comparision}.
Table~\ref{table:metrics_comparision} shows the comparisons of MAE and F-measure.
Note that $\max F_\beta$ is adopted by almost all methods except DEF~\cite{zhuge2019def}, which only reports $\mathrm{mean}\ F_\beta$.
We report the MAE, F-measure and S-measure of our method for a direct comparison.
Our ResNet based model achieves best results and consistently outperforms all other methods on all five datasets under different measurements, demonstrating the effectiveness of our proposed model.
Moreover, our VGG based model also ranks the top among VGG based methods.
This confirms that our proposed feature polishing method is effective and compatible with different backbone structures.
In Fig.~\ref{fig:PR_comparision}, we compare the PR curves and F-measure curves of different approaches on five datasets.
We can see that the PR curves of our method show better performance than others with a significant margin.
In addition, the F-measure curves of our method locate consistently higher than other methods.
This verifies the robustness of our method.

\textbf{Visual Comparison.}
Fig.~\ref{fig:visual_comparision} shows some example results of our model along with other six
state-of-the-art methods for visual comparisons.
We observe that our method gives superior results in complex backgrounds (row 1-2) and low contrast scenes (row 3-4).
And it recovers meticulous details (row 5-6, note the suspension cable of Golden Gate and the legs of the mantis).
From this comparison, we can see that our method performs robustly facing these challenges and produces better saliency maps.

\subsection{Ablation Study}
\textbf{Backbone.}
VGG-16~\cite{simonyan2014very} is a commonly used backbone by previous works~\cite{zhang2018bi,liu2018picanet}.
To demonstrate the capability of our proposed method to cooperate with different backbones, we introduce how it is applied to the multi-level features computed from VGG-16.
This adaption is straightforward.
VGG-16 contains 13 convolutional layers and 2 fully connected layers, along with 5 max-pooling layers which split the network into 6 blocks.
The 2 fully connected layers are first transformed to convolutional layers, and then the 6 blocks generate outputs with decreasing spatial resolutions, i.e. 256, 128, 64, 32, 16, 8, if the input image is set to the fixed size of 256x256.
These multi-level feature maps are fed into PFPN as described in Section~\ref{approach} to obtain the saliency map.
Table~\ref{table:metrics_comparision} shows the comparisons with other VGG based state-of-the-art methods and Table~\ref{table:res_ablation} shows the evaluations of various number of FPMs.
We can see that our method based on VGG-16 also shows excellent performance, which confirms that our method is effective for feature refining and generalizable to different backbones.

\begin{table}
    \tiny
	\setlength{\tabcolsep}{4pt}
	\centering
	 \caption{\small
		Ablation evaluations of PFPM with different $T$, the number of FPMs.
		The numbers in ($\cdot$) denote the value of $T$.
		PFPN-V denotes the models with VGG~\cite{simonyan2014very} as backbone.
		PFPN~(2 FPM)\ddag~denotes the FPMs the same share weights.
		Full metrices are given in supplementary materials.}
	\label{table:res_ablation}
	\begin{tabular}{l|cccc|cccc}
	\hline
	\multirow{2}*{Settings}&\multicolumn{4}{c|}{ECSSD}&\multicolumn{4}{c}{DUTS-TE}\\
	&MAE&max\ F&mean\ F&S  &MAE&max\ F&mean\ F&S\\
	\hline\hline
	PFPN~(0 FPM)     &0.048&0.928&0.894&0.911    &0.052&0.851&0.811&0.862    \\
	PFPN~(1 FPM)     &0.036&0.946&0.921&0.928    &0.040&0.884&0.848&0.883   \\
	PFPN~(2 FPM)\ddag&0.041&0.944&0.914&0.924    &0.043&0.876&0.840&0.884 \\
	PFPN~(2 FPM)     &0.033&0.949&0.926&0.932    &0.037&0.888&0.858&0.887  \\
	PFPN~(3 FPM)     &0.032&0.950&0.929&0.932    &0.037&0.888&0.862&0.889 \\
	\hline
	PFPN-V~(0 FPM)   &0.057&0.911&0.883&0.890    &0.058&0.825&0.793&0.837    \\
	PFPN-V~(1 FPM)   &0.045&0.931&0.905&0.908    &0.046&0.853&0.825&0.862   \\
	PFPN-V~(2 FPM)   &0.040&0.938&0.915&0.916    &0.042&0.868&0.836&0.864   \\
	PFPN-V~(3 FPM)   &0.040&0.939&0.915&0.920    &0.043&0.868&0.839&0.873  \\
	\hline
	\end{tabular}
 \end{table}

\textbf{Feature Polishing Module.}
To confirm the effectiveness of the proposed FPM, we conduct an ablation evaluation by varying the number of FPM employed.
The results with $T$ ranging from 0 to 3 on ECSSD and DUTS-TE are shown in Table~\ref{table:res_ablation}.
For $T=0$, two transition modules are directly connected without employing FPM, and for $T > 0$, FPM is applied $T$ times in between the two transition modules, as illustrated in Fig.~\ref{pipeline}.
Other settings, including the loss and training strategy, are kept the same for these evaluations.
For ResNet based models, we can see that FPM significantly boosts the performance than the plain baseline with no FPM, and the performances increase gradually with using more FPMs.
Actually the PFPN with 1 FPM and 2 FPMs both have great improvement progressively.
When it comes to $T=3$, the lift of accuracy converges and the improvement is marginal.
Similar phenomena can be observed with the VGG based PFPN.
This supports our argument that multiple FPMs progressively polish the representations so as to improve the final results.
We suppose the accuracy converges due to the limited scale of current dataset.
And we also conduct an experiment that a PFPN with 2 FPMs share the same weights.
The conclusion is that compared to PFPN (0 FPM), it has great improvement.
However, compared to PFPN (1 FPM) and PFPN (2 FPMs), the performance decay.
Although PFM can refine multi-level features, separate weights make FPM learning to refine features better according to different refinement stages.


 \begin{table}[hbt]
	\centering
	\caption{\small
		Quantitative comparison of the models with or without dense conditional random field~(DenseCRF) as a post-process.
		Full metrices are given in supplementary materials.}
	\small
	\renewcommand\tabcolsep{3pt}
	\label{table:crf_ablation}
	\begin{tabular}{l|cccc}
	\hline
	\multirow{2}*{Settings}&\multicolumn{4}{c}{DUTS-TE}\\
	&MAE&max\ F&mean\ F&S\\
	\hline\hline
	DSS~\cite{hou2017deeply}        &0.056&0.825&0.814&0.824 \\
	PiCA~\cite{liu2018picanet}      &0.051&0.860&0.816&0.868 \\
	PiCA+crf  						&0.041&0.866&0.855&0.862 \\
	PFPN                            &0.037&0.888&0.858&0.887 \\
	PFPN+crf                        &0.037&0.871&0.866&0.858 \\
	\hline
	\end{tabular}
 \end{table}

\textbf{DenseCRF.}
The dense connected conditional random field (DenseCRF~\cite{krahenbuhl2011efficient}) is widely used by many methods~\cite{hou2017deeply,liu2018picanet} as a post-process to refine the predicted results.
We investigate the effects of DenseCRF on our method.
The results are listed in Table~\ref{table:crf_ablation}.
DSS~\cite{hou2017deeply} reports the results with DenseCRF.
Both results with or without DenseCRF are reported for PiCA~\cite{liu2018picanet} and our method.
We can see that previous works can benefit from the long range pixel similarity prior brought by DenseCRF.
Furthermore, even without DenseCRF post-processing, our method performs better than other models with DenseCRF.
However, DenseCRF does not bring benefits for our method, where we find that DenseCRF only improves the MAE on a few datasets, but decreases the F-measure on all datasets.
This indicates that our method already sufficiently captures the information about the saliency objects from the data, so that heuristic prior fails to provide more help.

\subsection{Visualization of feature polishing}
\begin{figure}[hbt]
	\centering
	\setlength{\tabcolsep}{0.1mm}
	\renewcommand{\arraystretch}{0.5} 
	\begin{tabular}{cccccc}
	\includegraphics[width=13.6mm]{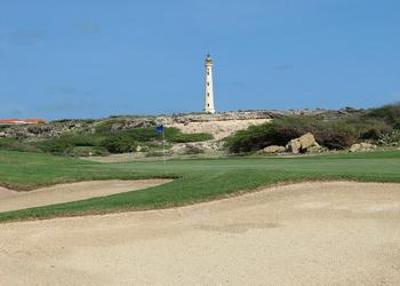}
	&\includegraphics[width=13.6mm]{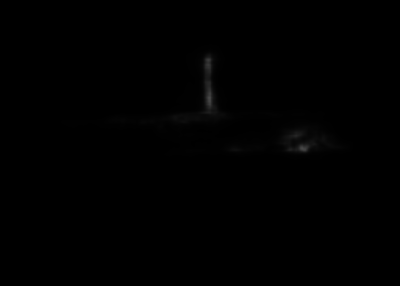}
	&\includegraphics[width=13.6mm]{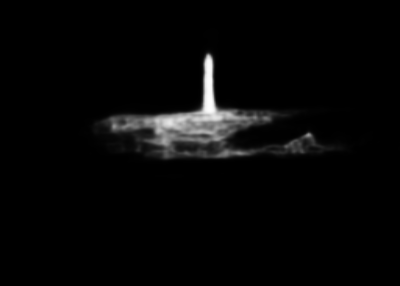}
	&\includegraphics[width=13.6mm]{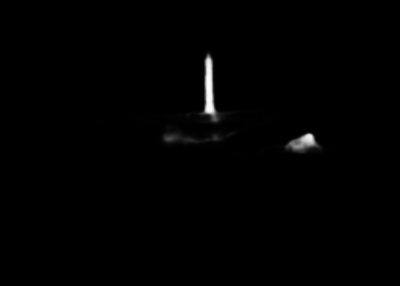}
	&\includegraphics[width=13.6mm]{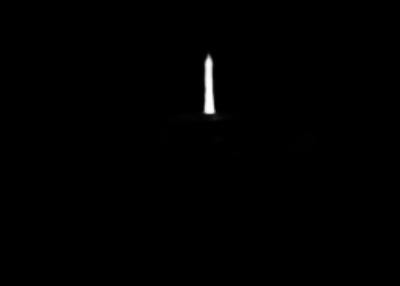}
	&\includegraphics[width=13.6mm]{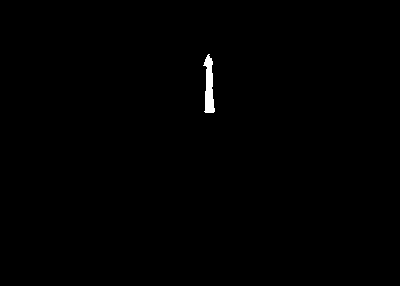}\\
 
	\includegraphics[width=13.6mm]{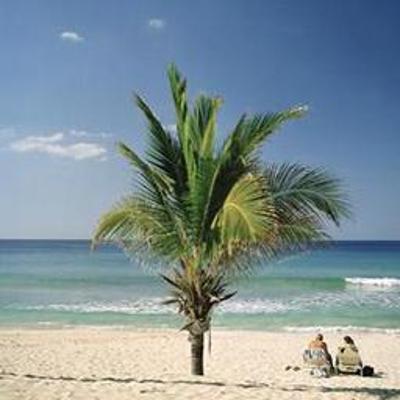}
	&\includegraphics[width=13.6mm]{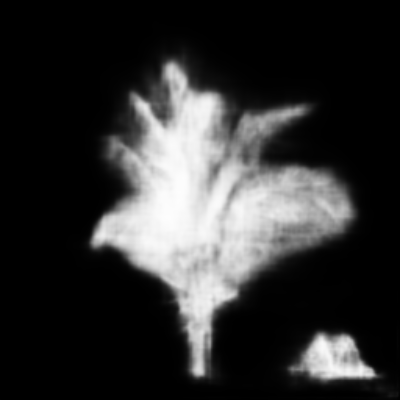}
	&\includegraphics[width=13.6mm]{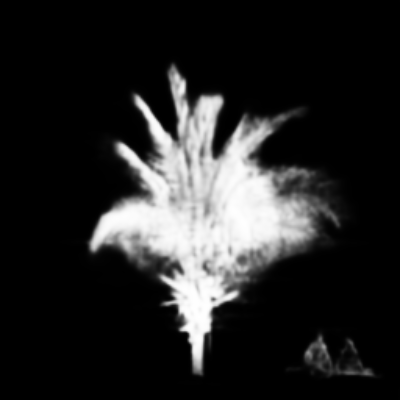}
	&\includegraphics[width=13.6mm]{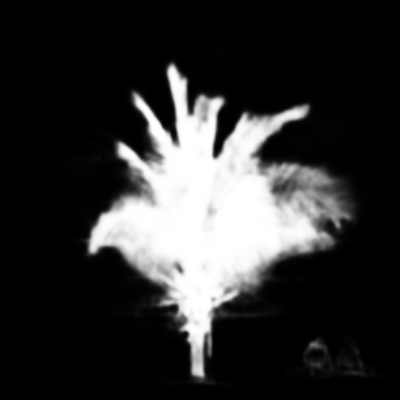}
	&\includegraphics[width=13.6mm]{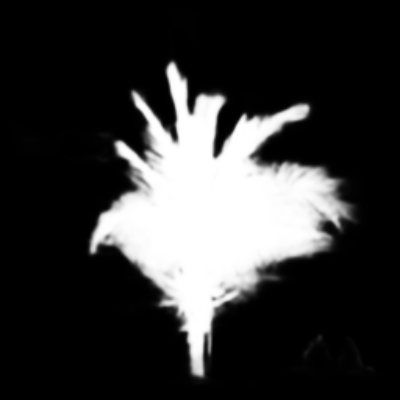}
	&\includegraphics[width=13.6mm]{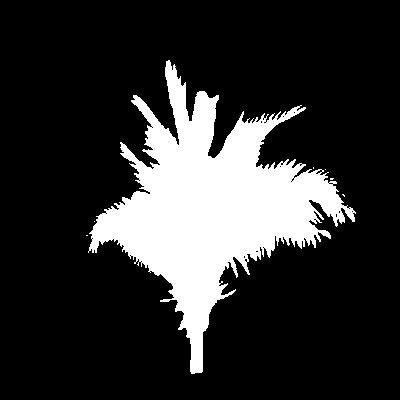}\\
 
	\includegraphics[width=13.6mm]{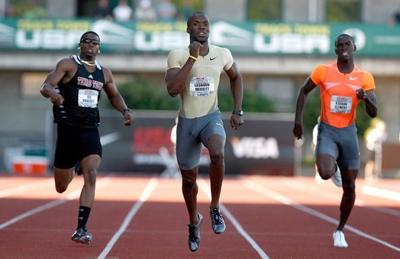}
	&\includegraphics[width=13.6mm]{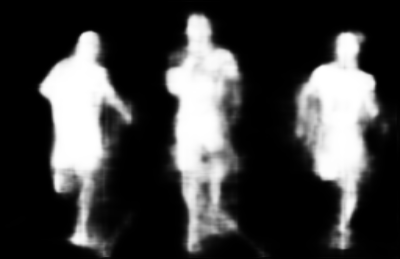}
	&\includegraphics[width=13.6mm]{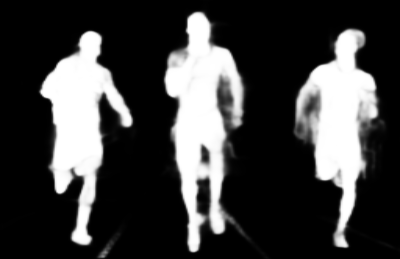}
	&\includegraphics[width=13.6mm]{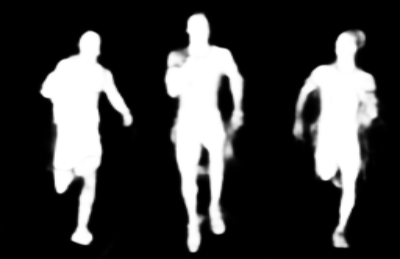}
	&\includegraphics[width=13.6mm]{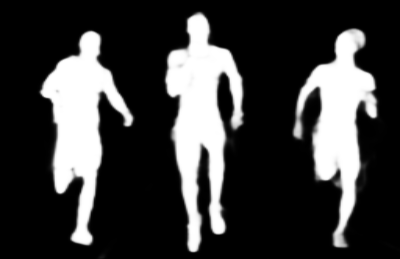}
	&\includegraphics[width=13.6mm]{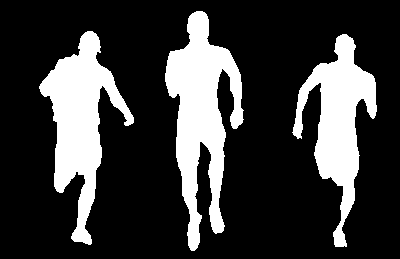}\\
 
 
 
	(a) & (b) & (c)&(d) & (e) & (f)\\
	
	\end{tabular}
 \caption{\small
	Saliency maps predicted by our proposed PFPN with various numbers of FPMs.
 (a) Original images. (f) Ground truth. (b)-(e)  Saliency maps predicted by PFPN with $T= 0\sim3$ FPMs, respectively.}
 \label{fig:pfm_vis}
 \end{figure}

In this section, we present an intuitive understanding of the procedure of feature polishing.
Since directly visualizing the intermediate features are not straightforward, we instead compare the results of our model with different numbers of FPMs.
Several example saliency maps are illustrated in Fig.~\ref{fig:pfp_results} and Fig.~\ref{fig:pfm_vis}.
We can see that the quality of predicted saliency maps is monotonically getting better with increasing number of FPMs, which is consistent with quantitative results in Table~\ref{table:res_ablation}.
Specifically, the model with $T=0$ can roughly detect the salient objects in the images,  which benefits from rich semantic information of multi-level feature maps. 
As more FPMs are employed, more details are recovered and cluttered results are eliminated.

\section{Conclusion}
We have presented a novel Progressive Feature Polishing Network for salient object detection.
PFPN focuses on improving the multi-level representations by progressively polishing the features in a recurrent manner.
For each polishing step, a Feature Polishing Module is designed to directly integrate high level semantic concepts to all lower level features, which reduces information loss.
Although the overall structure of PFPN is quite simple and tidy,
empirical evaluations show that our method significantly outperforms 16 state-of-the-art methods on five benchmark datasets under various evaluation metrics.

\section{Acknowledgement}
This work was supported by Alibaba Group through Alibaba Innovative Research Program.
Xiaogang Jin is supported by the Key Research and Development Program of Zhejiang Province (No. 2018C03055) and the National Natural Science Foundation of China (Grant Nos. 61972344, 61732015).

{\small
\bibliographystyle{aaai}
\bibliography{74.AAAI-WangB}
}

\end{document}